\documentclass[lettersize,journal]{IEEEtran}
\usepackage{amsmath,amsfonts}
\usepackage{algorithmic}
\usepackage{algorithm}
\usepackage{array}
\usepackage[caption=false,font=normalsize,labelfont=sf,textfont=sf]{subfig}
\usepackage{textcomp}
\usepackage{stfloats}
\usepackage{url}
\usepackage{verbatim}
\usepackage{graphicx}
\usepackage{cite}

\usepackage{multirow}  
\usepackage{tabularx} 
\usepackage{enumitem} 
\usepackage{caption}
\usepackage{booktabs}
\usepackage{tabularx}
\usepackage{siunitx}
\usepackage{makecell}
\usepackage[colorlinks=true, linkcolor=blue, citecolor=blue, filecolor=blue, urlcolor=blue]{hyperref}

\begin{document}

\title{Through the Looking Glass: A Dual Perspective on Weakly-Supervised Few-Shot Segmentation}

\author{Jiaqi Ma,
        Guo-Sen Xie,
        Fang Zhao,
        and Zechao Li,~\IEEEmembership{Senior Member,~IEEE}
\thanks{Received 25 August 2025; revised 9 March 2026 and 13 May 2026; accepted 22 June 2026.
This work was supported by the National Natural Science Foundation of China (Grant Nos. 62425603, 62276134, and 62476124) and the Basic Research Program of Jiangsu Province (Grant No. BK20240011). Corresponding author: Guo-Sen Xie.

Jiaqi Ma, Guo-Sen Xie, and Zechao Li are with the School of Computer Science and Engineering, Nanjing University of Science and Technology, Nanjing 210094, China. E-mail: machiachi@163.com, gsxiehm@gmail.com, zechao.li@njust.edu.cn.}
\thanks{Fang Zhao is with the School of Intelligence Science and Technology, Nanjing University, Suzhou 215163, China. E-mail: fzhao@nju.edu.cn.}
} 

\markboth{Journal of \LaTeX\ Class Files,~Vol.~14, No.~8, August~2021}%
{Shell \MakeLowercase{\textit{et al.}}: A Sample Article Using IEEEtran.cls for IEEE Journals}


\maketitle

\begin{abstract}
Meta-learning aims to uniformly sample homologous support-query pairs, characterized by the same categories and similar attributes, and extract useful inductive biases through identical network architectures. However, this identical network design results in over-semantic homogenization. To address this, we propose a novel homologous but heterogeneous network. By treating support-query pairs as dual perspectives, we introduce heterogeneous visual aggregation (HA) modules to enhance complementarity while preserving semantic commonality. To further reduce semantic noise and amplify the uniqueness of heterogeneous semantics, we design a heterogeneous transport (HT) module. Finally, we propose heterogeneous CLIP (HC) textual information to enhance the generalization capability of multimodal models. In the weakly-supervised few-shot semantic segmentation (WFSS) task, with only 1/24 of the parameters of existing state-of-the-art models, TLG achieves a 13.2\% improvement on Pascal-5\textsuperscript{i} and a 7.9\% improvement on COCO-20\textsuperscript{i}. To the best of our knowledge, TLG is also the first weakly-supervised (image-level) model that outperforms fully supervised (pixel-level) models under the same backbone architectures. The code is available at https://github.com/jarch-ma/TLG. 
\end{abstract}


\section{Introduction}

\begin{figure}[h]
\captionsetup{skip=0pt}
\vskip 0.1in
\begin{center}
\includegraphics[width=1\linewidth]{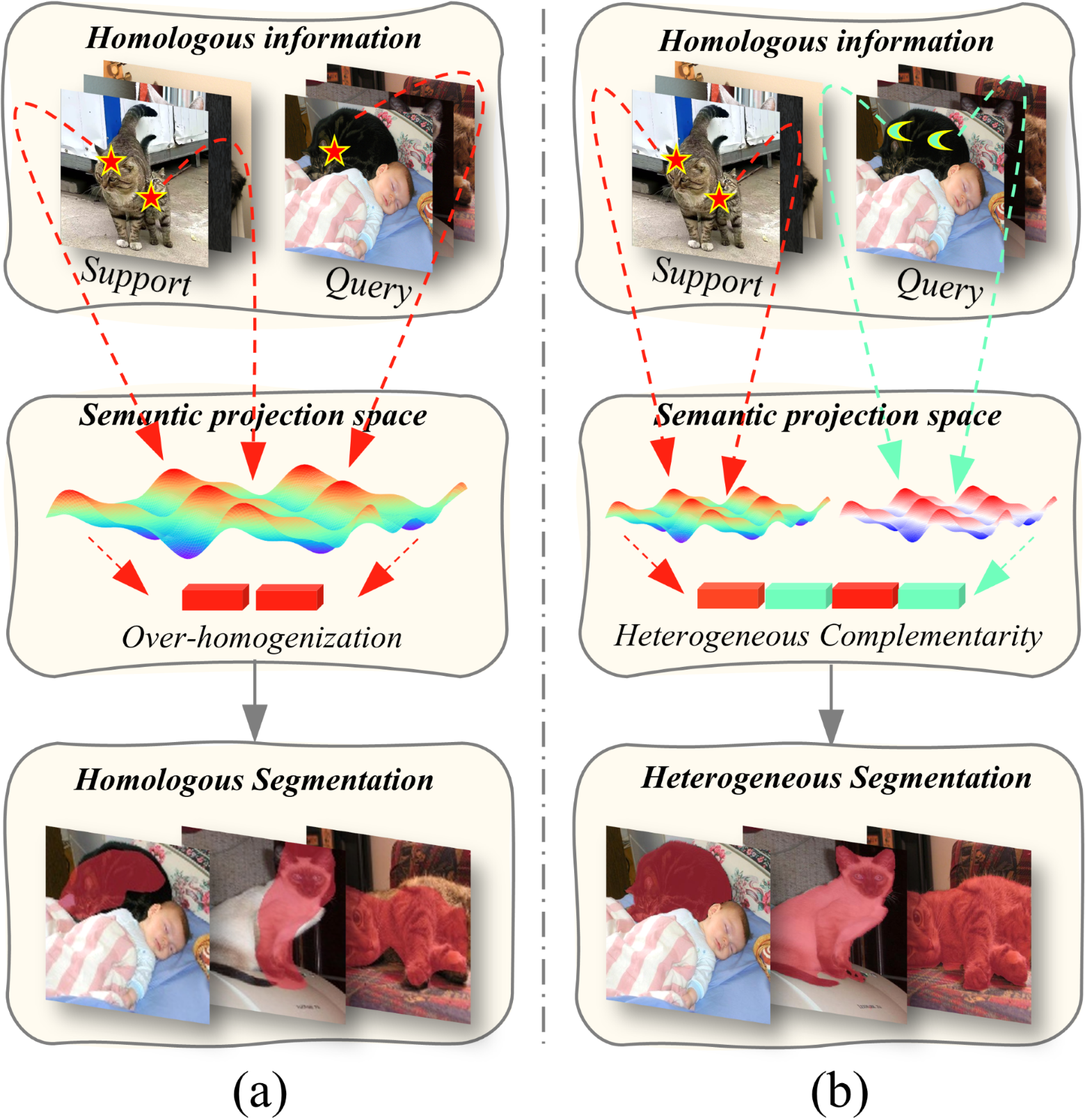}
\caption{(a) Traditional homologous network. Support-query pairs of the same category extract similar features and use identical architectures (Projection in the same semantic mapping space), resulting in over-homogenization of semantic features and limiting model performance. (b) Heterogeneous network. Support-query pairs extract distinct features and adopt different design patterns (Projection in the different semantic mapping space), enhancing semantic richness from multiple perspectives and promoting heterogeneous complementarity to improve model performance.}
\label{fig1}
\end{center}
\vskip -0.3in
\end{figure}
\vskip -0.05in

\IEEEPARstart{T}{h}e Meta-learning, owing to its strong adaptability, has attracted considerable attention for effectively addressing challenges such as data scarcity and long-tailed distributions in real-world tasks ~\cite{paper1, paper2, paper3, touvron2023llama, pmlr-v202-wang23x, hospedales2021meta,li2023knowledge,Peng_2019_ICCV}. The field has been extensively studied, leading to the emergence of various methods, such as similarity-based metric learning and prototype-based class-agnostic induction approaches ~\cite{zhang2024few, zhao2021mul,8403294, azad2024medical}. However, these methods fail to fully exploit the potential of meta-learning, as they employ identical network architectures for both the support and query branches, thereby overlooking their inherent semantic complementarity \cite{mm23slowfastego, GPT4Ego, dgz_tip_thinkmatter,li1,li3}.

In biology, homologous but heterogeneous describes structures sharing a common evolutionary origin (homology) but diverging into distinct morphological or functional traits (heterogeneity) \cite{jones2012genomic}. Similarly, meta-learning exhibits this property. For example, support and query targets may be a \textquoteleft \textit{rottweiler}\textquoteright{} and a \textquoteleft \textit{corgi}\textquoteright{}, both belonging to the coarse-grained category \textquoteleft \textit{dog}\textquoteright{}, reflecting semantic homology, while differing in fine-grained attributes such as size and coat color, illustrating instance-level heterogeneity. As shown in Figure \ref{fig1}(a), traditional meta-learning uses a single network and shared embedding space for support and query, causing the model to overemphasize shared features, such as \textquoteleft \textit{head}\textquoteright{}, and neglect heterogeneous details, leading to under-segmentation. In contrast, Figure \ref{fig1}(b) shows that using distinct but semantically aligned architectures introduces heterogeneous representations while preserving homology, enabling more effective semantic complementarity \cite{tang2023context, tang11299097, li2}. \textbf{T}hrough the \textbf{L}ooking \textbf{G}lass,\footnote{Inspired by the classic literary work \cite{carroll2007through}, where mirrored and real worlds oppose yet reflect each other, the metaphor parallels TLG’s design: heterogeneous branches sharing a homologous category.} we propose a novel homologous but heterogeneous network based on a dual-perspective approach, named \textbf{TLG}. TLG captures richer complementary semantic features, unlocking the full potential of meta-learning models.

\begin{figure*}[t]
		\includegraphics[width=1\linewidth]{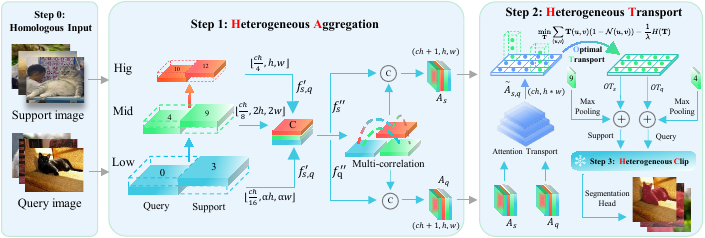}
		\caption{Overview of our TLG Framework. (1) Heterogeneous Aggregation (HA) module. To complement heterogeneous semantics based on shared semantics, intra-layer heterogeneity is leveraged (e.g., using layer 3 for support and layer 0 for query within the lower layers 0–3), along with inter-layer correlations. (2) Heterogeneous Transport (HT) module. The HT module enhances contextual relevance through attention transport, while suppressing noise via optimal transport, and subsequently enriches semantic diversity with heterogeneous residuals. (3) Heterogeneous CLIP (HC) module. By incorporating multimodal heterogeneous CLIP text information, the robustness of TLG is enhanced.}
        \label{fig2}
\vskip -0.15in
\end{figure*}

As shown in Figure \ref{fig2}, TLG consists of three modules: 1. Heterogeneous Aggregation (HA): support and query extract semantic embeddings from similar but slightly different backbone layers, maximizing heterogeneous semantic richness while preserving semantic commonality. 2. Heterogeneous  Transport (HT): Uses the optimal transport algorithm to reduce noise inevitably introduced during the aggregation of heterogeneous semantic information. 3. Heterogeneous Clip (HC): Leverages CLIP’s heterogeneous textual prior to enhance TLG’s multimodal representation and improve model generalization.

Our insight can be encapsulated in the phrase: \textit{Segmentation of the heterogeneous, by the heterogeneous, and for the heterogeneous.} \textbf{Of} refers to the segmentation of heterogeneous elements, defined by their diversity and complementarity. \textbf{By} indicates that the segmentation process is driven by heterogeneous feature embeddings. \textbf{For} signifies that the heterogeneous is not only used in segmentation but also serves as a novel design paradigm, advancing the development of meta-learning. To sum up, our contributions can be listed as:

\begin{itemize}
    \item We have identified the issue of homogeneity in the meta-learning paradigm and propose a novel homologous but heterogeneous network design paradigm. 
    \item To validate the effectiveness of the homologous but heterogeneous network, we propose a multimodal-based visual Heterogeneous Aggregation (HA) Module and Heterogeneous Transport (HT) Module, as well as a Heterogeneous CLIP (HC) Module based on Clip textual information.
    \item In the WFSS task, with only 1/24 of the parameters of existing state-of-the-art models, TLG achieves a 13.2\% improvement on Pascal-5\textsuperscript{i} and a 7.9\% improvement on COCO-20\textsuperscript{i}.
    \item To the best of our knowledge, TLG is the first weakly-supervised model (image-level labels) with the same backbone to achieve superior performance compared to existing fully-supervised (pixel-level labels) state-of-the-art models.
\end{itemize}

\section{Related Works}

\subsection{Meta Learning}
Traditional neural networks assume that training and test data are independent and identically distributed (i.i.d.) \cite{10938027,8935497, 10471355}. However, this assumption often deviates from real-world scenarios, leading to significant performance degradation when models encounter out-of-distribution domains ~\cite{hsu2020generalized, GUO2024106458, kim2026seg4diff, li2026exploring}. Meta-learning, or learning-to-learn, aims to improve the learning algorithm itself by leveraging knowledge from similar or related tasks, thus enabling generalization across multiple domains ~\cite{tang2022learning,zhao2021recent,10411850, wang2026opmapper}. This paradigm offers new solutions to address current challenges related to data and computational bottlenecks, as well as generalization issues.

However, current meta-learning is still limited by homogenized network design strategies, which constrain the potential of model performance \cite{Silva-Rodriguez_2024_CVPR, 10413635, 10516268}. Therefore, this paper proposes a new heterogeneous meta-learning design paradigm, aiming to overcome the limitations of existing meta-learning models and better address the complexities of real-world scenarios.

\subsection{Heterogeneous Network}
Heterogeneous refers to systems or populations that exhibit uniformity at a macro level, yet contain inherent diversity at a micro or individual level. This duality is often observed in ecosystems, cellular networks, and genetic populations, where shared traits or functions exist alongside diverse responses to environmental pressures ~\cite{jones2012genomic, lin2025tapb}. Correspondingly, in graph neural networks, this heterogeneity is manifested as different types of nodes and edges. Heterogeneous Graph Neural Networks (HGNNs) can use special mechanisms to distinguish the relationships between different types of nodes and edges, thereby enabling information propagation and learning in complex graph data ~\cite{zhang2019heterogeneous}. There are other heterogeneous network related to meta-learning, such as HetMAML ~\cite{ding2023cross}, which, through the meta-learning paradigm, learns both task-specific knowledge and globally shared knowledge when handling heterogeneous task distributions (HTD).

However, these studies focus on specific heterogeneous task distributions and merely adopt the meta-learning paradigm without introducing fundamental innovations. In contrast, this paper treats the support-query pairs during the data collection phase as naturally homologous data with the same category and similar attributes. In the network design phase, we apply differentiated processing to the support and query to accommodate the heterogeneous network, thereby overcoming the limitation of excessive homogeneity in the feature space of meta-learning.

\subsection{Weakly-Supervised Few-shot Segmentation}
Weakly-supervised few-shot semantic segmentation (WFSS) has garnered increasing attention in recent years due to its alignment with the complex and diverse characteristics of real-world application scenarios \cite{9303471, Tang_2019_ICCV, 9409690, Xie_2021_CVPR, Xie_2021_ICCV, 9362263, 10078892}. \cite{raza2019weakly} were the first to apply image-level labels to WFSS tasks, but these labels were only used during the testing phase, making it an incomplete WFSS model. \cite{siam2020weakly} reduced the distribution discrepancy between visual and textual modalities by leveraging multi-modal interaction information. ~\cite{lee2022pixel} determined the similarity between labels by measuring the distance between their word embeddings, but the accuracy remains limited. \cite{ying2021weakly} improved object-level representations by aggregating local object features. IMR-HSNet~\cite{wang2023iterative} first introduced the large visual-language model CLIP, refining pseudo-masks through iterative interactions between support and query images. MIAPNet \cite{pandey2023weakly} proposes a unified zero-shot and few-shot segmentation model, which significantly improves performance by learning context vectors through batch aggregation. Recently, AFANet \cite{ma2024afanet} transformed image RGB information into frequency domain distributions, incorporating more semantic information, and proposed an online-learning CLIP adapter, significantly boosting WFSS performance. However, the model has a large number of parameters and high training costs. In contrast, TLG employs a heterogeneous network structure, achieving superior performance with only 1/24 of the trainable parameters compared to AFANet.

\section{Methodology}

\subsection{Overview}

The core idea of TLG is \textit{seeking common ground while preserving differences} in support-query pairs to address feature space homogeneity during meta-learning, thereby enhancing the generalization of the model. As shown in Fig. \ref{fig2}, we developed three core modules based on a visual and textual multimodal scenario to validate the generalizability of the homologous but heterogeneous network design concept. These modules include: (1) Heterogeneous Aggregation (HA) module, which is designed for visual scenarios. It extracts heterogeneous information to mitigate semantic over-homogenization while simultaneously reducing model parameters and enhancing semantic complementarity. (2) The Heterogeneous Transport (HT) module, which establishes contextual correlations through an attention mechanism and mitigates the noise arising from heterogeneous feature aggregation by leveraging the Optimal Transport (OT) algorithm. (3) Heterogeneous CLIP (HC) module, which is specifically designed for textual scenarios. It establishes semantic associations between fine-grained foreground and co-occurring background elements to fully exploit the prior knowledge of CLIP, thereby enhancing the generalization capability of TLG.

\subsection{Preliminary}

In this study, we adopt a common meta-learning paradigm for 1-way K-shot weakly-supervised semantic segmentation. The standard approach is to construct a base dataset \(D_{\text{base}}\) and a novel dataset \(D_{\text{novel}}\), where \(D_{base}\cap D_{novel}=\phi\). \(D_{\text{base}}\) consists of seen classes and is further partitioned into a support set $\mathcal{S}$ and query set $\mathcal{Q}$ for meta-training across multiple episodes, while dataset \(D_{\text{novel}}\)  is used to assess the model's ability to generalize to unseen classes. It is defined as:
\begin{align}
  \mathcal{S}=\left\{\left(I_{s}^{k}, M_{s}^{k}\right)\right\}_{k=1}^{K}, \quad \mathcal{Q}=\left\{(I_{q}, M_{q}\right)\}.
\end{align}
Here,  ${I_{s}^{k}\in \mathbb{R}^{C \times H \times W}}$ and ${M_{s}^{k}\in \mathbb{R}^{H \times W}}$ represent the input images and their corresponding binary masks, respectively. However, in the WFSS setting, ground truth masks ${M_{s}^{k}}$ are unavailable, with only image-level labels provided. Thus, the primary task is to generate pseudo-masks $\tilde{M}_{s}^{k}$ and $\tilde{M}_{q}$ for the support \(\mathcal{S}\) and query $\mathcal{Q}$, respectively, using class activation maps (CAMs) \cite{zhou2016learning}.

\begin{align}
  \tilde{M}_{s}^{k}=\frac{{\text{relu}}({{t}_{i}^{k}}{W}_{s}^{k})}{\text{max}(\text{relu}({{t}_{i}^{k}}{W}_{s}^{k}))},
\end{align}
where, ${t}_{i}^{k}$ denotes the class features extracted by the CLIP text encoder, while ${W}_{s}^{k}$ denotes the embedding weights corresponding to given support image $I_{s}^{k}$. In this task, $K \in \{1, 5\}$ denotes the shot setting in meta-learning.

\subsection{Heterogeneous Aggregation (HA)}
AFANet extracts cross-granularity prior information from the backbone to enhance model representation. However, this cross-granularity standard is still limited to a homogenized feature space, where support and query share the same feature granularity. Here, we introduce a heterogeneous information extraction approach. As shown in Figure \ref{fig2}, given the input image support $I_{s}\in \mathbb{R}^{3 \times H \times W}$ and query $I_{q}\in \mathbb{R}^{3 \times H \times W}$, we follow the cross-granularity feature extraction method by extracting semantic embeddings from the backbone's low, middle, and high layers, respectively. To enhance the heterogeneity of semantic information and improve feature space complementarity, support is extracted from layers 3, 9, and 12, while query is extracted from layers 0, 4 and 10, respectively. This process can be described as:
 \begin{equation}
    \label{eq:equ1}
    f_{s}^{L}, f_{q}^{L} = F_{\theta}(X_{s}, X_{q}), \quad L \in \{\text{low}, \text{middle}, \text{high}\},
    \end{equation}
where \( F_{\theta} \) denotes the heterogeneous feature extraction process, while \( f_{s}^{L} \in \mathbb{R}^{ch \times H \times W} \) and \( f_{q}^{L} \in \mathbb{R}^{ch \times H \times W} \) encapsulate heterogeneous information from the low, middle, and high layers, where \( L_{\text{low}} \in \{0, 1, 2, 3\} \), \( L_{\text{middle}} \in \{4, 5, 6, 7, 8, 9\} \), and \( L_{\text{high}} \in \{10, 11, 12\} \).

After feature extraction, to achieve effective heterogeneous information aggregation, the support and query images must be aligned. During this process, the channel depth of the heterogeneous information in the low, middle, and high layers is reduced by a scaling factor of $1 / 2^{\alpha}$, where $\alpha \in {4, 2, 1}$, while the corresponding spatial dimensions are upsampled to a fixed size, as follows:

\begin{equation}
    \label{eq:equ2}
    f_{s, q}^{'} = \text{proj}(f_{s}^{L}, f_{q}^{L}), \quad L \in \{\text{low}\},
\end{equation}
\begin{equation}
    \label{eq:equ3}
    f_{s, q}^{'} = \text{upsample}(\text{proj}(f_{s}^{L}, f_{q}^{L})), \quad L \in \{\text{middle}, \text{high}\},
\end{equation}
 \begin{equation}
    \label{eq:equ4}
     f_{s, q}^{''} = \sum_{l=1}^{L} f_{s, q}^{'}, \quad L \in \{\text{low}, \text{middle}, \text{high}\},
    \end{equation}
where, $proj$ denotes spatial depth adjustment  using a 1x1 convolution kernel, while upsample refers to bilinear interpolation, with spatial dimensions uniformly fixed at (50, 50). $f_{s, q}^{''}$represents the preliminary heterogeneous aggregation. To standardize the notation, let $f_{s, q}^{'}$ denote the union of $f_{s}^{'}$ and $f_{q}^{'}$, with $f_{s, q}^{''}$ defined analogously.

The purpose of heterogeneity is complementarity, not differentiation, so it is also important to aggregate the commonalities between the semantic spaces by
{\small
\begin{equation}
    \label{eq:equ5}
     A_{s}  = \text{Concat}(f_{s}^{''} + \text{Init}, \text{Corr}(f_{s}^{'})),
    \end{equation}
}
\vspace{-0.1in} 
{\small
\begin{equation}
\label{eq:equ6}
     A_{q}  = \text{Concat}(f_{q}^{''}, \text{Corr}(f_{q}^{'})).
\end{equation}
}
$Init$ represents a Gaussian-distributed initial value, denoted as QUERY. $Corr$ is a 4D convolution \cite{zhang2022hsnet} used to capture multi-layer correlations.

\subsection{Heterogeneous Transport (HT)}
After the HA module, the semantic features complement each other with heterogeneous information while retaining commonalities. However, this inevitably introduces some semantic noise, leading to redundancy. Therefore, the HT module utilizes cross-attention \cite{NIPS2017_3f5ee243} to transform CNN convolutional features into attention, endowing them with contextual awareness to highlight the semantics of interest.
\begin{equation}
\label{eq:equ7}
\tilde{A}_{s,q} = \text{Softmax}\left( \frac{A_{s,q}^{''}  W (A_{s,q}^{'} )^T}{\sqrt{d_k}} \right) A_{s,q}^{'} W
\end{equation}

The feature $A_{s,q}^{'} \in \mathbb{R}^{ch \times HW}$ denotes the aggregated feature after flattening, while $A_{s,q}^{''} \in \mathbb{R}^{ch \times HW}$ represents the attention weights learned by $A_{s,q}^{'} \in \mathbb{R}^{ch \times HW}$ through self-attention, where $W$ denotes the learnable weights.

Thus, the attention mechanism can more clearly distinguish the semantics of the target object from noise interference\cite{an2025generalized, 8240654, 10.3389/fphar.2021.814858, deng2026training}. To effectively remove semantic noise, TLG models this problem as an optimal transport (OT) problem. Concretely, we apply the Sinkhorn algorithm \cite{cuturi2013sinkhorn, bunne2024optimal, zheng2024adapting} to achieve optimal pixel assignment by minimizing the transport cost, effectively removing noisy features. The optimization formula can be represented as:
{\small
\begin{equation}
\label{eq:equ8}
OT_{s,q} =\min_{\tau \in \gamma } \sum_{i,j}{\tau(i,j)(1-C_{s,q}^{f}(i,j))} - \frac{1}{\lambda }H(\tau ),
\end{equation}
}
\vskip -0.15in
{\small
\begin{equation}
\label{eq:equ9}
H(\tau ) = - \sum_{i,j} \tau(i,j) \log \tau(i,j),
\end{equation}
} where, $C_{s,q}^{f}(i,j) \in \mathbb{R}^{H \times W}$ denotes the heterogeneous foreground attention, obtained by thresholding ($>0$) the heterogeneous features aggregated in the HA module, where the foreground attention approximately represents the similarity between foreground regions. The transport cost matrix is defined as \( 1 - C_{s,q}^{f}(i,j) \), where large values of \(C_{s,q}^{f}(i,j)\) indicate high semantic similarity to the foreground, leading to lower transport costs and prioritized alignment in the optimal transport process. In contrast, smaller values correspond to low-similarity regions with higher transport costs, which are treated as semantic noise and suppressed during optimization. $H(\tau)$ is the entropy regularization term that stabilizes the solution and improves distribution.

To avoid excessive semantic alignment and enhance the diversity of feature representations, we introduce heterogeneous residuals (HR), which extract 1/3 of the original input features to preserve complementary information and increase representation heterogeneity between inputs.
{\small
\begin{equation}
\label{eq:equ10}
OT_{s}^{'} = OT_{s} + MaxPool(Select(f_{s}^{'})),
\end{equation}
}
\vskip -0.15in
{\small
\begin{equation}
\label{eq:equ11}
OT_{q}^{'} = OT_{q} + MaxPool(Select(f_{q}^{'})).
\end{equation}
}
$Select$ denotes selecting a single layer from the low-, mid-, and high-level feature stages. In our implementation, the support branch extracts features from the 9th layer, while the query branch extracts features from the 4th layer. For the selected features, max pooling is applied to both the support and query branches.

\begin{figure}[t]
    \centering
    \includegraphics[width=1\linewidth]{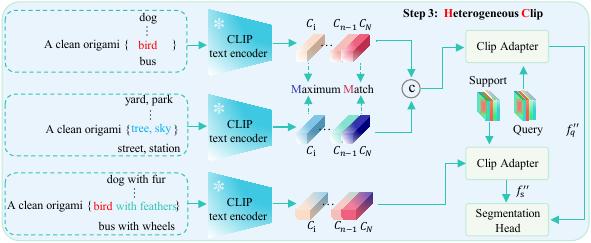}
    \caption{The HC module comprises three CLIP text encoders. From top to bottom, the coarse-grained foreground and background encoders suppress irrelevant background interference through maximum likelihood category matching. They then complement the fine-grained encoder and visual semantics to achieve multimodal heterogeneous information integration.}
    \label{fig3}
    \vskip -0.2in
\end{figure}

\subsection{Heterogeneous CLIP (HC)}

Purely visual information may struggle to capture diverse semantic details, particularly in complex scenes and ambiguous backgrounds. In this section, we incorporate multimodal CLIP textual heterogeneous information to improve the model's robustness and generalization ability.

In this study, we utilize CLIP text prompts and pseudo masks based on CLIP-ES \cite{lin2023clip}. However, CLIP-ES’s background prompts are not specifically associated with foreground categories, instead mapping all background semantic priors to a common space, leading to association ambiguity. For instance, when the foreground category is \textit{cat}, background prompts like \textit{sky} and \textit{ocea}n make it difficult for the model to effectively distinguish the foreground and background.

\begin{figure}[t]
    \centering
    \includegraphics[width=1\linewidth]{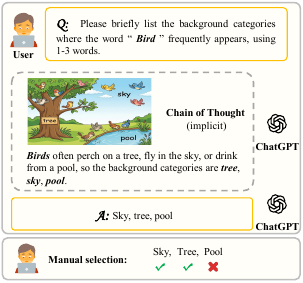}
    \caption{Visualization of foreground co-occurrence background category generation. The chain of thought is provided solely to facilitate understanding, and ChatGPT’s reasoning process is implicit.}
    \label{gpt}
\end{figure}

As shown in Figure \ref{fig3}, we propose the maximum matching CLIP prompt. Specifically, we first identify the foreground categories in the dataset and then retrieve the co-occurring backgrounds based on these. In this process (Figure \ref{gpt}), we use ChatGPT to obtain the two backgrounds most likely associated with each foreground, for example, ‘bird’ with ‘tree’ and ‘sky’. Additionally, to enhance the heterogeneous information in the CLIP text priors, we introduce fine-grained prompts tailored to foreground attributes, such as ‘\textit{aeroplane with wings}’, ‘\textit{bird with feathers}’. This approach facilitates more precise modeling of foreground characteristics, thereby improving the alignment between textual and visual features. Finally, to enhance the generalization capability of TLG, we adopt a heterogeneous architectural approach for the online CLIP Adapter in AFANet, which can be expressed as:
\begin{equation}
\label{eq:equ12}
\overline{M}_{s} = Adapter(T _{s}^{'},  \overline{T }_{gain} ),
\end{equation}
\begin{equation}
\label{eq:equ13}
\overline{M}_{q} = Adapter(T _{q}^{'},  Concat(\overline{T}_{fg},\overline{T}_{bg})).
\end{equation}
$\overline{M}_{s,q} \in \mathbb{R}^{ch \times H\times W}$ denotes the multimodal information obtained after merging visual and textual semantics. $\overline{T}_{gain}\in \mathbb{R}^{C_{N} \times hw}$ is fine-grained textual information. Where, $C_{N} = N$ represents the number of categories, either Pascal-5\textsuperscript{i} (N=20) or COCO-20\textsuperscript{i} (N=80). $\overline{T}_{fg}\in \mathbb{R}^{C_{N} \times hw}$ and $\overline{T}_{bg}\in \mathbb{R}^{C_{N} \times hw}$ represent the foreground and background textual information, respectively.

\begin{table*}[t]
\centering
\vskip -0.15in
\caption{Comparisons of regular and weakly-supervised few-shot semantic segmentation methods on Pascal-5\textsuperscript{i}. \textit{P} and \textit{I} denote pixel-level labels and image-level labels, respectively. The \textcolor{red}{red} and \textcolor{blue}{blue} colors respectively represent the optimal and suboptimal results. ${\color{red} \uparrow }$ represents the value of improvement from the SOTA model (AFANet).}
\resizebox{\textwidth}{!}{ 
\begin{tabular}{cc|l|c|p{1cm}cccc|p{1cm}cccc}

\toprule
                            & \multirow{3}{*}{Backbone} & \multicolumn{1}{c|}{\multirow{3}{*}{Methods}} & \multirow{3}{*}{A. Type} & \multicolumn{5}{c}{1-shot} & \multicolumn{5}{c}{5-shot} \\
\cmidrule(lr){5-14}
                            & & & & \centering $5^{0}$ & $5^{1}$ & $5^{2}$ & $5^{3}$ & Mean & \centering $5^{0}$ & $5^{1}$ & $5^{2}$ & $5^{3}$ & Mean \\
\cmidrule(lr){1-14}
                            & \multirow{6}{*}{VGG16} 
                            & HDMNet \cite{peng2023hierarchical}                               & \textit{P} & \centering 64.8 & 71.4 & 67.7 & 56.4 & 65.1 & \centering 68.1 & 73.1 & 71.8 & 64.0 & 69.3 \\
                            & & PFENet++ \cite{luo2023pfenet++}                        & \textit{P} & \centering 59.2 & 69.6 & 66.8 & 60.7 & 64.1 & \centering 64.3 & 72.0 & 70.0 & 62.7 & 67.3 \\
                            & & DRNet \cite{chang2024drnet}                              & \textit{P} & \centering 65.4 & 66.2 & 54.2 & 53.8 & 59.9 & \centering 68.9 & 68.2 & 64.6 & 57.1 & 64.7 \\
\cmidrule(lr){3-14}  
                            & & Pix-MetaNet \cite{lee2022pixel}                      & \textit{I} & \centering 36.5 & 51.7 & 45.9 & 35.6 & 42.4 & \centering - & - & - & - & - \\
                            & & IMR-HSNet \cite{wang2023iterative}                       & \textit{I} & \centering 58.2 & 63.9 & 52.9 & 51.2 & 56.5 & \centering 60.5 & 65.3 & 55 & 51.7 & 58.1 \\
                            & & AFANet \cite{ma2024afanet}                                           & \textit{I} & \centering \textcolor{blue}{64.1} & \textcolor{blue}{65.9} & \textcolor{blue}{59.2} & \textcolor{blue}{56.9} & \textcolor{blue}{61.5} & \centering \textcolor{blue}{66.3} & \textcolor{blue}{68.7} & \textcolor{blue}{61.6} & \textcolor{blue}{59.7} & \textcolor{blue}{64.1} \\
                            & & \textbf{TLG (Ours)}                                        & \textit{I} & \centering \textcolor{red}{72.5} & \textcolor{red}{81.5} & \textcolor{red}{77.5} & \textcolor{red}{72.8} & ${\color{red} 76.1^{\uparrow 14.6}}$ & \centering \textcolor{red}{73.0} & \textcolor{red}{82.4} & \textcolor{red}{78.1} & \textcolor{red}{73.7} & ${\color{red} 76.8^{\uparrow 12.7}}$ \\
\cmidrule(lr){2-14}  
                            & \multirow{10}{*}{Resnet50} 
                            & HDMNet \cite{peng2023hierarchical}                              & \textit{P} & \centering 71.0 & 75.4 & 68.9 & 62.1 & 69.4 & \centering 71.3 & 76.2 & 71.3 & 68.5 & 71.8 \\
                            & & PFENet++ \cite{luo2023pfenet++}                        & \textit{P} & \centering 60.6 & 70.3 & 65.6 & 60.3 & 64.2 & \centering 65.2 & 73.6 & 74.1 & 65.3 & 69.6 \\
                            & & DRNet \cite{chang2024drnet}                              & \textit{P} & \centering 66.1 & 68.8 & 61.3 & 58.2 & 63.6 & \centering 69.2 & 73.9 & 65.4 & 65.3 & 68.5 \\
                            & & PGMA-Net \cite{zhang2023rpmg}                             & \textit{P} & \centering 73.4 & 80.8 & 70.5 & 71.7 & 74.1 & \centering 74.0 & 81.5 & 71.9 & 73.3 & 75.2 \\
                            & & VRP-SAM \cite{sun2024vrp}                             & \textit{P} & \centering 73.9 & 78.3 & 70.6 & 65.0 & 71.9 & \centering - & - & - & - & - \\
\cmidrule(lr){3-14}
                            & & Siam et al. \cite{siam2020weakly}                     & \textit{I} & \centering 49.5 & 65.5 & 50.0 & 49.2 & 53.5 & \centering - & - & - & - & - \\
                            & & Zhang et al. \cite{zhang2022weakly}                           & \textit{I} & \centering 56.9 & 62.5 & 60.3 & 49.9 & 57.4 & \centering - & - & - & - & - \\
                            & & IMR-HSNet \cite{wang2023iterative}                       & \textit{I} & \centering 62.6 & \textcolor{blue}{69.1} & 56.1 & 56.7 & 61.0 & \centering 63.6 & 69.6 & 56.3 & 57.4 & 61.8 \\
                            & & MIAPNet \cite{pandey2023weakly}                           & \textit{I} & \centering 41.7 & 51.3 & 42.2 & 41.8 & 44.2 & \centering - & - & - & - & - \\
                            & & AFANet \cite{ma2024afanet}                                         & \textit{I} & \centering \textcolor{blue}{65.7} & 68.5 & \textcolor{blue}{60.6} & \textcolor{blue}{61.5} & \textcolor{blue}{64.0} & \centering \textcolor{blue}{69.0} & \textcolor{blue}{70.4} & \textcolor{blue}{61.3} & \textcolor{blue}{64.0} & \textcolor{blue}{66.2} \\
                            & & \textbf{TLG (Ours)}                                        & \textit{I} & \centering \textcolor{red}{75.4} & \textcolor{red}{81.8} & \textcolor{red}{77.4} & \textcolor{red}{74.0} & ${\color{red} 77.2^{\uparrow 13.2}}$ & \centering \textcolor{red}{75.9} & \textcolor{red}{82.6} & \textcolor{red}{77.7} & \textcolor{red}{75.3} & ${\color{red} 77.9^{\uparrow 11.7}}$ \\
\bottomrule
\end{tabular}}
\label{table1}
\end{table*}

\begin{table*}[t]
\centering
\begin{center}
\begin{small}
\caption{Comparisons of regular and weakly-supervised few-shot semantic segmentation methods on COCO-20\textsuperscript{i}. \textit{P} and \textit{I} denote pixel-level labels and image-level labels, respectively. The \textcolor{red}{red} and \textcolor{blue}{blue} colors respectively represent the optimal and suboptimal results. ${\color{red} \uparrow }$ represents the value of improvement from the SOTA model (AFANet).}
\resizebox{\textwidth}{!}{ 
\begin{tabular}{cc|l|c|p{1cm}cccc|p{1cm}cccc}

\toprule
                            & \multirow{3}{*}{Backbone} & \multicolumn{1}{c|}{\multirow{3}{*}{Methods}} & \multirow{3}{*}{A. Type} & \multicolumn{5}{c}{1-shot} & \multicolumn{5}{c}{5-shot} \\
\cmidrule(lr){5-14}
                            & & & & \centering $5^{0}$ & $5^{1}$ & $5^{2}$ & $5^{3}$ & Mean & \centering $5^{0}$ & $5^{1}$ & $5^{2}$ & $5^{3}$ & Mean \\
\cmidrule(lr){1-14}
                            & \multirow{6}{*}{VGG16} 
                            & HDMNet \cite{peng2023hierarchical}                               & \textit{P} & \centering 40.7 & 50.6 & 48.2 & 44.0 & 45.9 & \centering 47.0 & 56.5 & 54.1 & 51.9 & 52.4 \\
                            & & PFENet++ \cite{luo2023pfenet++}                        & \textit{P} & \centering 38.6 & 43.1 & 40.0 & 39.5 & 40.3 & \centering 38.9 & 46.0 & 44.2 & 44.1 & 43.3 \\
                            & & DRNet \cite{chang2024drnet}                              & \textit{P} & \centering 33.5 & 33.4 & 32.1 & 34.2 & 33.3 & \centering 43.1 & 46.8 & 40.9 & 40.9 & 42.9 \\
\cmidrule(lr){3-14}  
                            & & IMR-HSNet \cite{wang2023iterative}                       & \textit{I} & \centering 34.9 & 38.8 & 37.0 & 40.1 & 37.7 & \centering 34.8 & 41.0 & 37.2 & 39.7 & 38.2 \\
                            & & AFANet \cite{ma2024afanet}                                          & \textit{I} & \centering \textcolor{blue}{38.3} & \textcolor{blue}{42.5} & \textcolor{blue}{42.9} & \textcolor{blue}{41.5} & \textcolor{blue}{41.3} & \centering \textcolor{blue}{37.9} & \textcolor{blue}{42.7} & \textcolor{blue}{40.6} & \textcolor{blue}{43.1} & \textcolor{blue}{41.1} \\
                            & & \textbf{TLG (Ours)}                                        & \textit{I} & \centering \textcolor{red}{46.7} & \textcolor{red}{55.4} & \textcolor{red}{51.0} & \textcolor{red}{50.2} & ${\color{red} 50.8^{\uparrow 9.5}}$ & \centering \textcolor{red}{46.8} & \textcolor{red}{55.0} & \textcolor{red}{50.8} & \textcolor{red}{49.1} & ${\color{red} 50.4^{\uparrow 9.3}}$ \\
\cmidrule(lr){2-14}  
                            & \multirow{10}{*}{Resnet50} 
                            & HDMNet \cite{peng2023hierarchical}                               & \textit{P} & \centering 43.8 & 55.3 & 51.6 & 49.4 & 50.0 & \centering 50.6 & 61.6 & 55.7 & 56.0 & 56.0 \\
                            & & PFENet++ \cite{luo2023pfenet++}                        & \textit{P} & \centering 40.9 & 44.8 & 39.7 & 38.8 & 41.0 & \centering 45.7 & 52.4 & 49.1 & 47.2 & 48.6 \\
                            & & DRNet \cite{chang2024drnet}                              & \textit{P} & \centering 42.1 & 42.8 & 42.7 & 41.3 & 42.2 & \centering 47.7 & 51.7 & 47.0 & 49.3 & 49.0 \\
                            & & VRP-SAM \cite{sun2024vrp}                             & \textit{P} & \centering 48.1 & 55.8 & 60.0 & 51.6 & 53.9 & \centering - & - & - & - & - \\
\cmidrule(lr){3-14}
                            & & Siam et al. \cite{siam2020weakly}                     & \textit{I} & \centering - & - & - & - & 15.0 & \centering - & - & - & - & 15.6 \\
                            & & Zhang et al. \cite{zhang2022weakly}                           & \textit{I} & \centering 33.3 & 32.0 & 29.2 & 29.2 & 30.9 & \centering - & - & - & - & - \\
                            & & IMR-HSNet \cite{wang2023iterative}                       & \textit{I} & \centering 39.5 & \textcolor{blue}{43.8} & 42.4 & 44.1 & 42.4 & \centering 40.7 & 46.0 & 45.0 & 46.0 & 44.4 \\
                            & & MIAPNet \cite{pandey2023weakly}                           & \textit{I} & \centering 34.9 & 23.4 & 12.4 & 18.3 & 22.2 & \centering - & - & - & - & - \\
                            & & AFANet \cite{ma2024afanet}                                         & \textit{I} & \centering \textcolor{blue}{40.2} & 45.1 & \textcolor{blue}{44.0} & \textcolor{blue}{45.1} & \textcolor{blue}{43.6} & \centering \textcolor{blue}{41.0} & \textcolor{blue}{49.5} & \textcolor{blue}{43.0} & \textcolor{blue}{46.9} & \textcolor{blue}{45.1} \\
                            & & \textbf{TLG (Ours)}                                        & \textit{I} & \centering \textcolor{red}{47.7} & \textcolor{red}{56.3} & \textcolor{red}{51.3} & \textcolor{red}{50.8} & ${\color{red} 51.5^{\uparrow 7.9}}$ & \centering \textcolor{red}{47.7} & \textcolor{red}{56.6} & \textcolor{red}{51.0} & \textcolor{red}{51.1} & ${\color{red} 51.6^{\uparrow 6.5}}$ \\
\bottomrule
\end{tabular}}
\label{table2}
\end{small} 
\end{center}
\vskip -0.15in
\end{table*}

\subsection{Loss Function}
In this study, TLG utilizes Binary Cross-Entropy (BCE) loss to optimize the pixel-wise discrepancy between the predicted results and pseudo-labels, thereby enhancing the accuracy and stability of the segmentation model. To further strengthen the heterogeneous difference between support and query, different hyperparameters are employed to optimize the impact of the loss. Therefore, the total loss of TLG is:
\begin{equation}
\label{eq:equ14}
\mathcal{L}_{\mathrm{BCE}}=-\sum_{i=1}^{\mathcal{C}} y_{i} \log \left(p_{i}\right),
\end{equation}
\vskip -0.2in
\begin{equation}
\label{eq:equ15}
\mathcal{L}_{\mathrm{all}}= \alpha \cdot \mathcal{L}_{\mathrm{BCE}}(\hat{M}_{s} ,\tilde{{M_{s} } } ) +  \beta  \cdot \mathcal{L}_{\mathrm{BCE}}(\hat{M}_{q} ,\tilde{{M_{q} } } ).
\end{equation}
Where, $\mathcal{L}_{\mathrm{BCE}}$ denotes the Binary Cross-Entropy loss, $\mathcal{C}$ is the number of classes, $y_{i}$ represents the pseudo-label classes, and $p_{i}$ is the predicted probabilities. $\hat{M}_{s,q}$ and $\tilde{M}_{s,q}$ correspond to the predicted mask and pseudo-mask, respectively. $\alpha$ and $\beta$ are hyperparameters.

\section{Experiments}
\subsection{Datasets}
To evaluate the segmentation and generalization performance of TLG, we conducted extensive experiments under a weakly-supervised few-shot setting on the Pascal-5\textsuperscript{i} and COCO-20\textsuperscript{i} datasets, using image-level labels for all datasets. Specifically, the datasets were divided into four folds, with three folds used for training and one fold for testing. Notably, each fold of Pascal-5\textsuperscript{i} contains five categories, whereas, due to the large size of COCO-20\textsuperscript{i}, each fold contains twenty categories.

\subsection{Implementation Details}
To ensure a fair comparison, we follow the standard WFSS setup, setting all image sizes to 400×400 and using VGG16 and ResNet50 as feature extraction backbones. In the experiments, we uniformly apply the AdamW optimizer with a weight decay coefficient of 1e-4, training for 80 epochs. The difference lies in the training configurations: for Pascal-5\textsuperscript{i}, the batch size is 16 with a learning rate of 4e-4, while for COCO-20\textsuperscript{i}, the batch size is 32 with a learning rate of 1e-4. During 5-shot testing, we randomly sample 1000 episodes for meta-testing on Pascal-5\textsuperscript{i}. Due to the large size of COCO-20\textsuperscript{i}, 1000 episodes are insufficient to accurately assess the robustness of TLG, so we adjust it to a more rigorous 5000 episodes. All experiments are conducted with two RTX 3090 GPUs.

\begin{figure*}[t]
    \centering
    \includegraphics[width=1\linewidth]{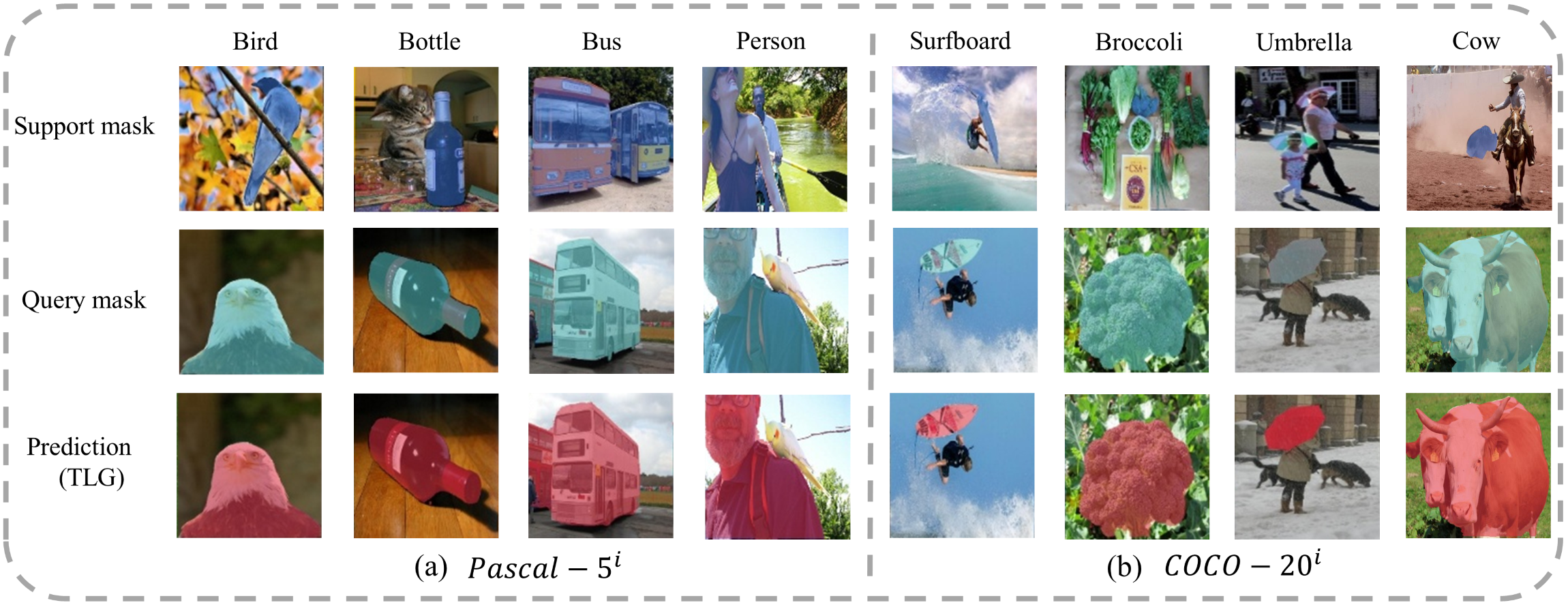}
    \caption{Qualitative Visualization Analysis: Visualizing Segmentation Results under a 1-Shot Setting on the Pascal-5\textsuperscript{i} and COCO-20\textsuperscript{i} Datasets.}
    \label{fig4}
    \vskip -0.2in
\end{figure*}

\subsection{Comparison with State-of-the-art}
In this section, we quantitatively and qualitatively compare TLG with the current state-of-the-art WFSS and FSS models under the same task settings.

$\mathbf{Pascal-5\textsuperscript{i}.}$ As shown in Table \ref{table1}, we compare the performance of TLG with current SOTA models under 1-shot and 5-shot settings, using ResNet50 and VGG16 as backbones. It is clearly observed that TLG demonstrates overwhelming performance advantages, regardless of the backbone or shot settings. With VGG16 as the backbone, TLG achieves a 4-fold average mIoU of 76.1 in the 1-shot setting, outperforming the current SOTA model AFANet by a significant margin of 14.6\%. In the 5-shot setting, the performance reaches 76.8, outperforming AFANet by 12.7\%. When using ResNet50 as the backbone, TLG achieves the highest performance, with mIoU scores of 77.2 and 77.9 in the 1-shot and 5-shot settings, respectively.

$\mathbf{COCO-20\textsuperscript{i}.}$ As shown in Table \ref{table2}, on the challenging COCO-20\textsuperscript{i} dataset, which contains a large number of complex scenes and diverse object categories, TLG consistently outperforms the current state-of-the-art models across different backbone and shot settings. With VGG16 as the backbone, TLG achieves mIoU scores of 50.8 (9.5\% improvement) and 50.4 (9.3\% improvement) under the 1-shot and 5-shot settings, respectively, showing substantial gains. With ResNet50, TLG attains its best performance, reaching 51.5 and 51.6 mIoU in the 1-shot and 5-shot settings, respectively.

\begin{table}[h]
\caption{Ablation study on various modules of TLG. The first row represents using only homologous features extracted by Backbone ResNet50. Following the same settings as AFANet \cite{ma2024afanet}, we set the support and query to layers 3, 9, and 12.}
\label{table3}
\centering
\small
\begin{tabularx}{\columnwidth}{
>{\centering\arraybackslash}p{0.7cm}
>{\centering\arraybackslash}p{0.7cm}
>{\centering\arraybackslash}p{0.7cm}
| *{4}{>{\centering\arraybackslash}X}
>{\centering\arraybackslash}c
}
\toprule
HA & HT & HC & $5^{0}$ & $5^{1}$ & $5^{2}$ & $5^{3}$ & Mean \\
\midrule
   &    &                  & 68.5   & 75.4  & 56.4   & 65.3   &  66.4 \\
\checkmark & &             & 73.2 & 78.4 & 71.3 & 73.7 & 74.2$^{\textcolor{red}{\uparrow 7.8}}$ \\
\checkmark & \checkmark &  & 74.3 & 81.9 & 74.1 & 74.2 & 76.1$^{\textcolor{blue}{\uparrow 1.9}}$ \\
\checkmark &  & \checkmark & 74.5 & 81.1 & 76.7 & 73.9 & 76.6$^{\textcolor{blue}{\uparrow 2.4}}$ \\
\checkmark & \checkmark & \checkmark & 75.4 & 81.8 & 77.4 & 74.0 & 77.2$^{\textcolor{red}{\uparrow 10.8}}$ \\
\bottomrule
\end{tabularx}

\vspace{0.1cm}

\raggedright Here, $\textcolor{red}{\uparrow}$ indicates the improvement over using only the backbone, while $\textcolor{blue}{\uparrow}$ denotes the respective performance gains of the HT and HC modules compared with the HA module.
\vskip -0.2in
\end{table}

\subsection{Heterogeneous layer selection strategy}

In this section, we explore the strategy for selecting heterogeneous features. As shown in Table \ref{table4}, we first used the fully connected features from layers 0 to 12 of both the support and query as the base, achieving an IoU of 74.8. Subsequently, we optimized the model using heterogeneous features. When the support features were derived from layers {0, 4, 10} and the query features from layers {3, 9, 12}, the IoU slightly decreased to 74.7. Notably, the features used in this setup accounted for only one-fourth of the fully connected features. However, the performance of TLG remained largely unchanged, which further indicates that homologous features lead to excessive semantic homogenization, thereby affecting the model's performance.

\begin{table}[h]
\caption{Ablation study on heterogeneous support and query layers. The backbone and dataset are ResNet50 and Pascal-5\textsuperscript{i}, respectively.}
\label{table4}
\centering
\small
\begin{tabularx}{\columnwidth}{
>{\centering\arraybackslash}X
>{\centering\arraybackslash}X
>{\centering\arraybackslash}X
}
\toprule
Support layers & Query layers & IOU\\  
\midrule
0-12      & 0-12      & 74.8 \\
0, 4, 10 & 3, 9, 12 & 74.7 \\
\textbf{3, 9, 12} & \textbf{0, 4, 10} & \textbf{75.4} \\
3, 9, 12 & 2, 7, 11 & 75.0 \\
\bottomrule
\end{tabularx}

\vspace{0.1cm}
\raggedright Note, 0–12 denotes the fully connected layer.
\vspace{-0.2cm}

\end{table}

Conversely, when the roles are reversed—support features drawn from {3, 9, 12} and query features from {0, 4, 10}—the IoU reaches its peak at 75.4. This indicates that high-level features, with their larger receptive fields, are more suitable as support features to provide semantic guidance, whereas low-level features, retaining finer-grained details, are better suited as query features for precise localization. Based on this intuition, we argue that the core of heterogeneous features lies in maximizing the differentiation between support and query features. For comparison, when the disparity is reduced—keeping support features at {3, 9, 12} but using intermediate layers {2, 7, 11} for queries—the IoU drops to 75.0. This trend is consistent with the heterogeneous feature visualizations in Figure \ref{fig10} and Figure \ref{fig11}: when support and query features originate from adjacent layers, semantic redundancy may arise, whereas greater layer disparity enhances heterogeneity and complementarity, thereby improving model performance.

\begin{table*}[ht]
\centering
\caption{Model comparison with computational metrics.}
\label{tab:model_comparison}
\scalebox{1}{
\begin{tabular}{@{}l
                l
                S[table-format=3.2]
                S[table-format=3.2]
                S[table-format=3.3]
                S[table-format=4.3]
                S[table-format=2.2]
                S[table-format=3.2]@{}}
\toprule
Model & Backbone & 
\multicolumn{1}{c}{\makecell{Learnable Param}} & 
\multicolumn{1}{c}{\makecell{Total Param}} & 
\multicolumn{1}{c}{\makecell{FLOPs (1-shot)}} & 
\multicolumn{1}{c}{\makecell{FLOPs (5-shot)}} & 
\multicolumn{1}{c}{\makecell{Inference Latency\\(1-shot)}} & 
\multicolumn{1}{c}{\makecell{Inference Latency\\(5-shot)}} \\
\midrule
\multirow{2}{*}{\makecell{AFANet  (baseline)}} & ResNet50  & 108.35 & 136.96 & 139.36 &  696.82 &  73.21 & 350.82 \\
                                      & VGG16     & 112.87 & 127.58 & 202.64 & 1013.00 &  57.92 & 289.57 \\
\midrule                                      
\multirow{2}{*}{\makecell{\textbf{TLG   (ours)}}} & ResNet50  &  4.47 &  28.03 &  41.25 &  206.24 &  36.72 & 174.38 \\
                               & VGG16     &  3.94 &  18.66 & 134.36 &  671.79 &  32.24 & 158.85 \\
\bottomrule
\multicolumn{8}{l}{\makebox[0pt][l]{\footnotesize Note: Param is measured in M, FLOPs in G, and Inference Latency in ms.}} \\
\end{tabular}
\label{table5}
}
\end{table*}

\subsection{Quantitative ablation analysis}
In this subsection, we present a detailed analysis of the various modules of TLG on the Pascal-5\textsuperscript{i} dataset. 

$\textbf{Effects of different modules}.$ As demonstrated in Table \ref{table3}, TLG exhibits progressive performance gains in ablation studies across heterogeneous modules. With the HA module alone, TLG fuses heterogeneous information to mitigate excessive homogeneity and achieve semantic complementarity, attaining a mean mIoU of 74.2\%. The subsequent introduction of the HT module strengthens contextual correlations through attention mechanisms, accentuates shared features across homologous sources, and eliminates noisy semantics via optimal transport theory, yielding a 1.9\% performance improvement. We also evaluated adding the HC module individually after the HA module. This configuration benefits from CLIP’s textual priors and the maximal matching mechanism between heterogeneous foreground and co-occurring background, achieving a 2.4\% improvement. Ultimately, integrating all modules allows the multimodal heterogeneous information to enhance TLG’s complex scene understanding via semantic complementarity in visual and textual spaces. This integration achieves a 10.8\% improvement over the backbone.

\begin{figure}[h]
    \centering
    \includegraphics[width=0.9\linewidth]{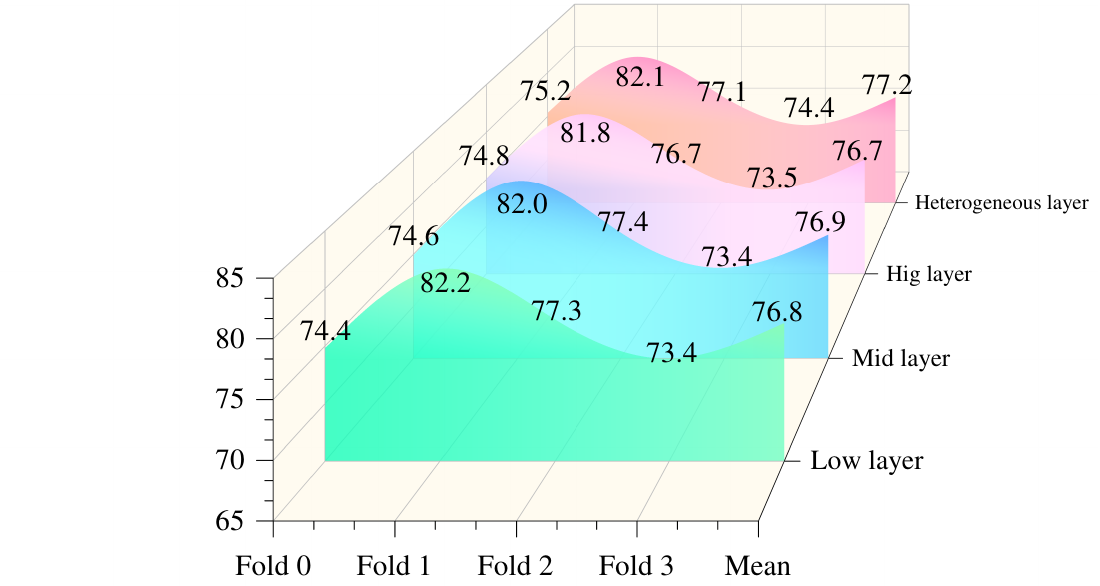}
    \caption{Ablation study between homologous but heterogeneous features.}
    \label{fig5}
    
\end{figure}

$\textbf{Effects of heterogeneous layers.}$ As shown in Figure \ref{fig5}, we analyzed the heterogeneous information extracted from different layers of the HA module. The pyramid ResNet50 backbone consists of 0–12 layers, categorized into three fine-grained levels based on feature channel depth and semantic spatial scale: layers 0–3, 4–9, and 10–12. For instance, layer 0 is a low layer relative to layer 3, and so forth. The semantic granularity extracted at each layer serves as the heterogeneous semantic feature.

In this ablation study, we first compare the extraction of homologous information from identical layers for both support and query. For example, both support and query extract features from layers 3, 9, and 12, or from layers 0, 4, and 10. The model's performance shows minimal change, attributed to the excessive similarity of semantic features between them. In contrast, the heterogeneous ablation experiments adopt a dual-branch perspective strategy. Specifically, support and query extract layers 3, 9 and 12, and layers 0, 4 and 10, respectively. While support's layer 3 and query's layer 0 belong to the same fine-grained level, providing common ground, the slight differences in granularity enhance the focus on distinct semantic features, fostering differentiation. The visual differences can be found in Figures \ref{fig10} and \ref{fig11}. This complementary interaction between heterogeneous support-query pairs significantly improves the generalization capability of the TLG model.

\begin{figure}[h]
    \centering
    \includegraphics[width=0.85\linewidth]{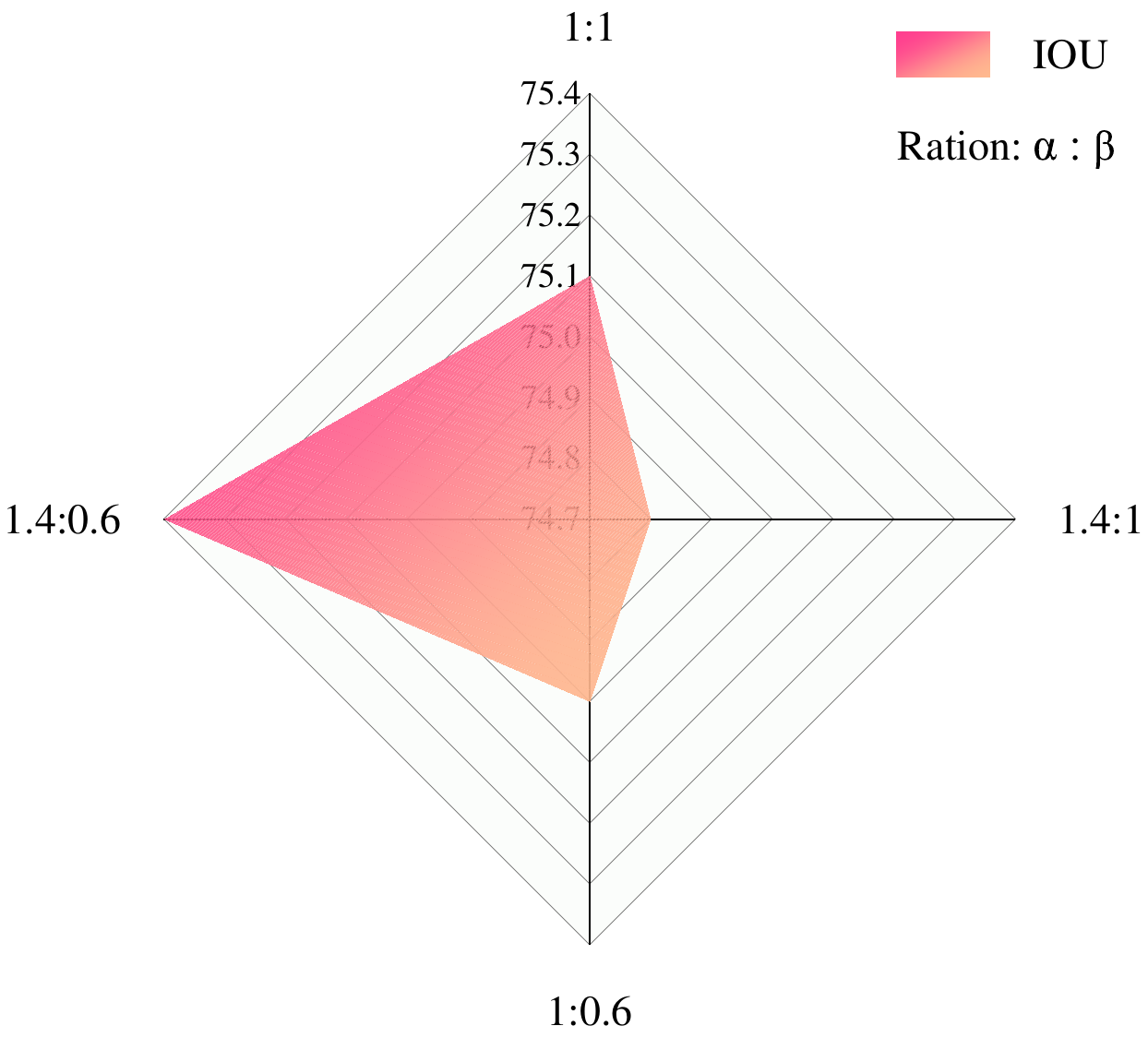}
    \caption{Ablation study of loss function hyperparameters. $\alpha$ and $\beta$ are the hyperparameters for the support and query, respectively.}
    \label{fig7}
    \vskip -0.15in
\end{figure}

\subsection{Effects of different loss hyperparameters}
In the experiment, TLG computes losses for both the support and query.To emphasize heterogeneity, hyperparameters $\alpha$ and $\beta$ are used for control. As illustrated in Figure \ref{fig7}, when both $\alpha$ and $\beta$ are set to 1, a balanced state is achieved, resulting in a performance of 75.1. When $\alpha$ and $\beta$ are set to 1, with one hyperparameter adjusted to 0.6 or 1.4, the performance slightly decreases to 75.0 and 74.8, respectively. However, when the heterogeneous loss becomes complementary—specifically, when $\alpha$ is 1.4 and $\beta$ is 0.6 the performance reaches its peak at 75.4. This also indicates that, during training, slightly increasing the support weights that provide semantic guidance as category prototypes helps improve segmentation performance on query images.

\begin{figure*}[t]
    \centering
    \includegraphics[width=1\linewidth]{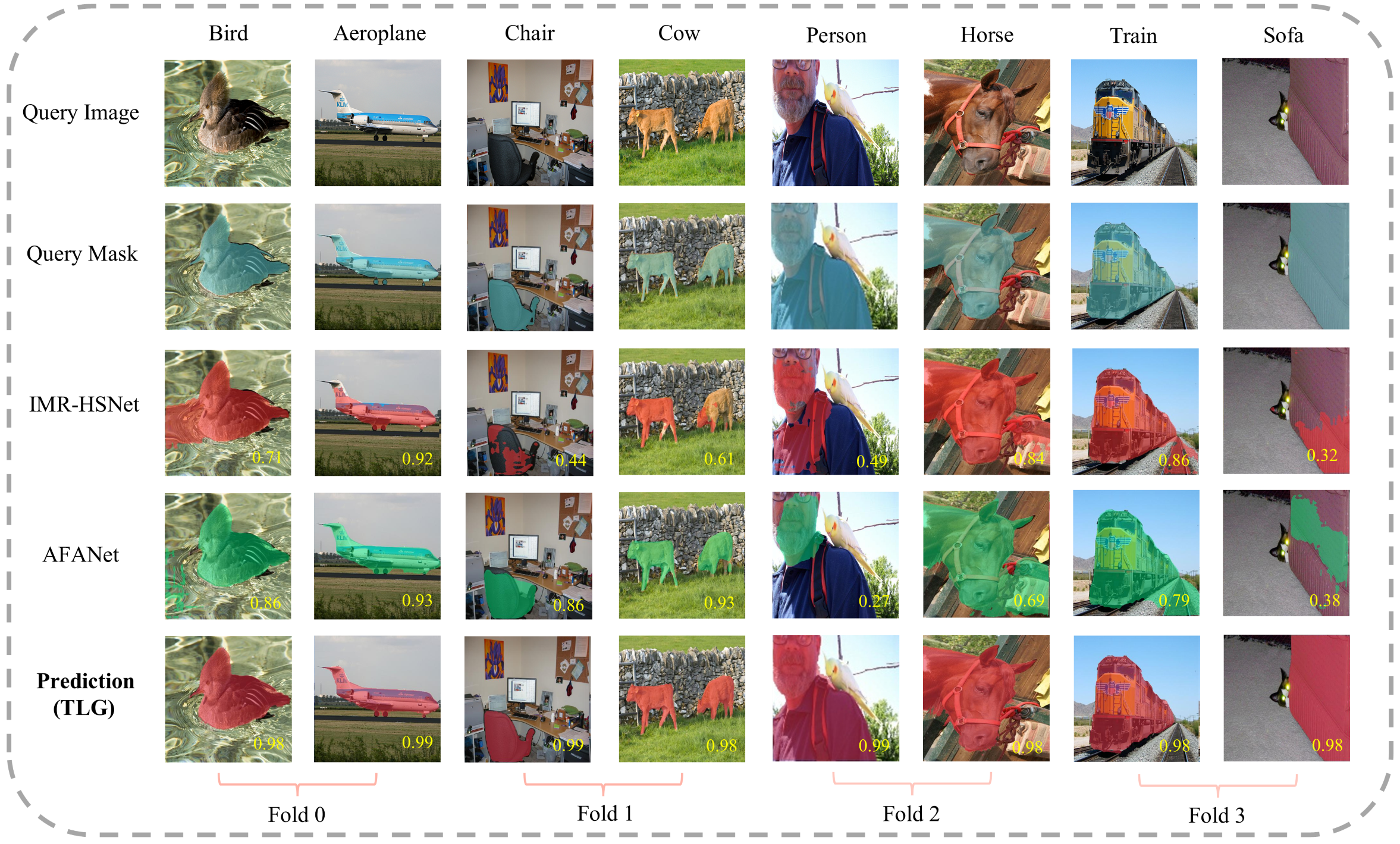}
    \caption{Comparative Qualitative Visualization Analysis with State-of-the-Art Models: Visualizing Segmentation Results under a 1-Shot Setting on the Pascal-5i Datasets. Each pair of columns corresponds to a fold in the meta-learning paradigm. The yellow numbers in the bottom right denote the Intersection over Union (IoU) scores.}
    \label{fig8}
    \vskip -0.1in
\end{figure*}

\begin{table}[h]
\centering
\caption{Ablation study on heterogeneous residuals.}
\label{pooling}
\resizebox{0.75\linewidth}{!}{%
\begin{tabular}{ccc}
\toprule
Support & Query & IoU (\%) \\
\midrule
-- & -- & 75.1 \\
average pooling & average pooling & 74.8 \\
max pooling & max pooling & \textbf{75.6} \\
max pooling & average pooling & 75.3 \\
average pooling & max pooling & 75.2 \\
\bottomrule
\end{tabular}%
}
\end{table}

\subsection{Effects of heterogeneous residuals}

Table~\ref{pooling} presents the IoU results of ablation studies on different heterogeneous residuals strategies for support and query features. This ablation study involves only the HA and HT modules. It can be observed that applying max pooling to both branches achieves the best performance, with an IoU of 75.6\%. The heterogeneity is mainly achieved through cross-layer selection of one-third of the original features, and the results demonstrate that the proposed residual design effectively enhances feature diversity and improves feature representation.

\subsection{Qualitative visualization analysis}

$\textbf{TLG segmentation result visualization.}$
In this section, we conduct a qualitative analysis of TLG's segmentation performance by selecting four categories each from the Pascal-5\textsuperscript{i} and COCO-20\textsuperscript{i} datasets (Figure \ref{fig4}), respectively. These eight categories encompass a range of typical scenarios, from dynamic organisms to static objects, and from regular shapes to complex textures, reflecting the diversity of semantic categories and structural features in real-world scenes. From left to right, in the first and fourth columns, the objects are a bird and a person (with bird interference). TLG accurately distinguishes and segments them, effectively capturing their semantic features. Similarly, in the third column (\textit{bus}), sixth column (\textit{broccoli}), and eighth column (\textit{cow}), the segmented objects face challenges such as partial occlusion, similar backgrounds, and object overlap. TLG successfully handles these complexities, demonstrating its robustness in intricate segmentation tasks.

\begin{figure*}[t]
    \centering
    \includegraphics[width=1\linewidth]{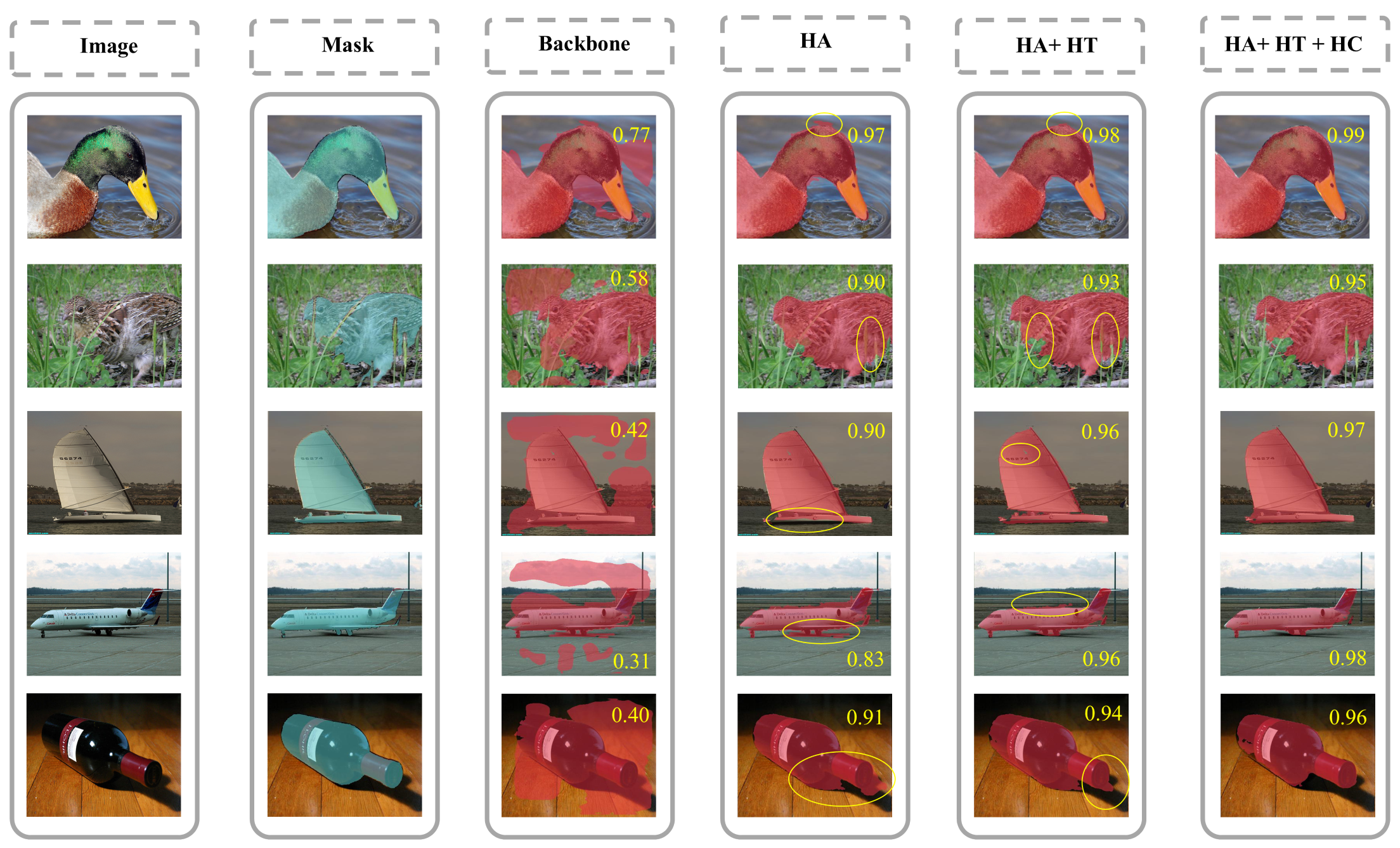}
    \caption{Ablation Study via Qualitative Visual Analysis: Visualizing Segmentation Results under a 1-Shot Setting on the Pascal-5i fold 0 Datasets. From left to right, the first two columns illustrate the raw input image and its corresponding ground-truth mask. “Backbone” denotes features extracted solely from ResNet-50 for segmentation visualization. “HA” refers to the Heterogeneous Aggregation module, “HT” denotes the Heterogeneous Transport module, and “HC” represents the Heterogeneous CLIP module, respectively. The yellow numbers in the top right denote the Intersection over Union (IoU) scores. The yellow circles highlight regions of over-segmentation or under-segmentation.}
    \label{fig9}
    \vskip -0.1in
\end{figure*}

$\textbf{Visualization comparison with SOTA methods.}$ To rigorously validate the robustness of TLG, we conducted comparative experiments with state-of-the-art models IMR-HSNet \cite{wang2023iterative} and AFANet \cite{ma2024afanet} on representative segmentation tasks. As illustrated in Figure \ref{fig8}, for co-occurring categories like \textit{bird} (Column 1) and \textit{train} (Column 7), conventional models erroneously segmented background elements (e.g., rails/water) due to feature entanglement, whereas TLG achieved precise semantic localization through heterogeneous feature decoupling. In boundary-ambiguous scenarios (Column 2, \textit{airplane}), IMR-HSNet suffered from under-segmentation due to limited receptive fields, while AFANet exhibited over-segmentation correlated with excessive parameters (for more details, see Figure \ref{fig12}). TLG attained boundary precision of 0.99 IoU via multi-modal heterogeneous semantic comprehension. For partially visible objects like \textit{sofa} (Column 8), conventional models showed limited performance (IoU$<$0.4) under single-modal constraints, whereas TLG achieved 0.98 IoU through cross-modal semantic reasoning. These results demonstrate that by integrating heterogeneous information, TLG not only enriches semantic expression but also enhances the model’s semantic reasoning ability, significantly improving segmentation accuracy and robustness.

\begin{figure*}[t!]
    \centering
    \vskip 0.1in
    \includegraphics[width=1\linewidth]{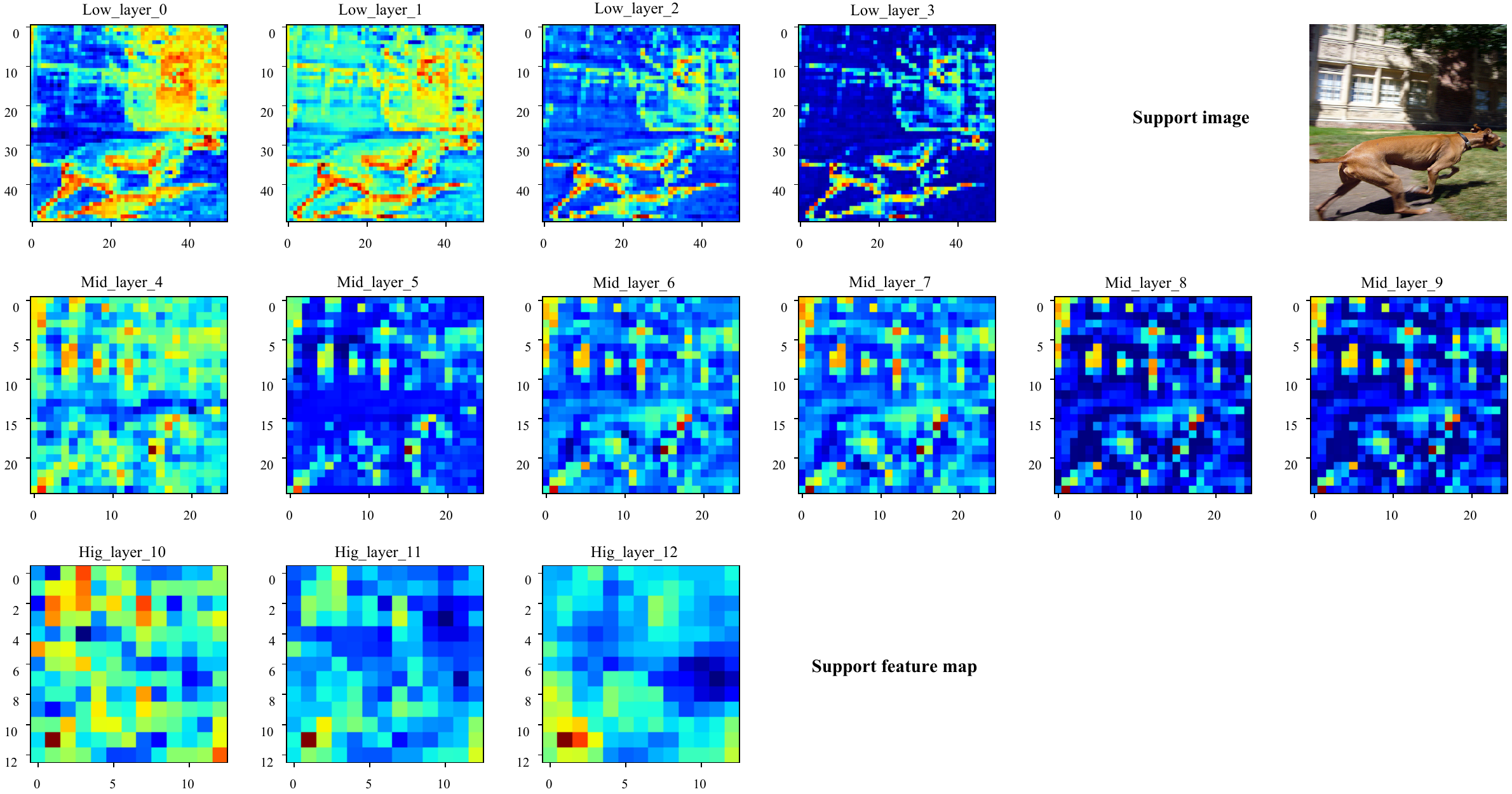}
    \caption{Visualization of heterogeneous features at different layers for support image. The first row corresponds to the low layer, while the second and third rows represent the middle layer and high layer, respectively. Relevant content is omitted below.}
    \label{fig10}
    \vskip 0.3in
    \includegraphics[width=1\linewidth]{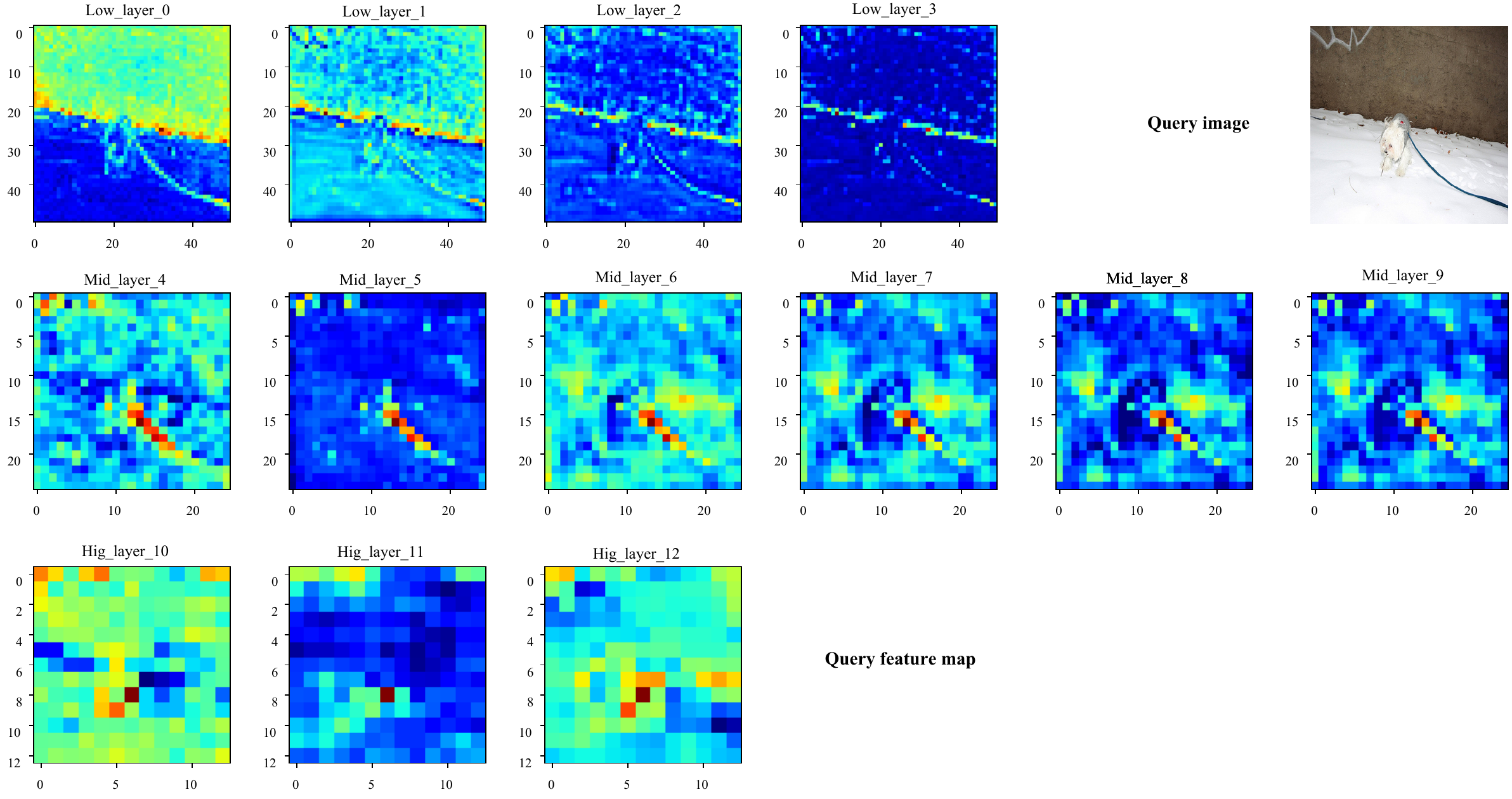}
    \caption{Visualization of heterogeneous features at different layers for query image. }
    \label{fig11}
\end{figure*}

$\textbf{TLG qualitative ablation analysis.}$ In this section, we present a segmentation visualization ablation study on the various modules of TLG. As shown in Figure \ref{fig9}, each column sequentially illustrates the segmentation outcomes of different module combinations. In the third column, the baseline model (Backbone) employs only ResNet50 to extract homologous features (using the AFANet framework), yielding a semantic understanding at a coarse level and insufficient foreground-background separation (mean IoU = 0.4). Upon integrating the Heterogeneous Aggregation (HA) module, the model leverages complementary heterogeneous features to significantly improve foreground recognition accuracy (mean IoU = 0.9), although the highlighted yellow regions in the fourth column indicate localized over-segmentation. Further incorporation of the Heterogeneous Transport (HT) module—utilizing a context-attention mechanism based on optimal transport theory—effectively suppresses feature noise and enables state-of-the-art boundary localization. Ultimately, by fusing prior knowledge from the base model via the Heterogeneous CLIP (HC) module, TLG achieves hierarchical semantic enhancement and precise reconstruction. Experimental results demonstrate that the cascade optimization mechanism of "heterogeneous aggregation—noise suppression—knowledge fusion" collectively underpins TLG’s superior performance.

\subsection{Heterogeneous feature visualization}
\label{F}
As shown in Figures \ref{fig10} and \ref{fig11}, we visualize the feature maps extracted at different backbone layers for both the support and query images. Layers 0–3 represent the low layer features, layers 4–9 correspond to the middle layer features , and layers 10–12 represent the high layer features. It can be observed that the backbone's lower layers focus more on shallow features, such as the texture of target objects, while the higher layers emphasize high-level semantic information. The mid-level layers, serving as an intermediate representation, strike a balance between these two extremes and are thus more commonly utilized for model training.

Furthermore, it can be observed that even within the same level (e.g., low, mid, or high layers), the semantic information captured by individual layers exhibits subtle differences. In meta-learning, extracting identical features from support and query images and using the same network structure for training can lead to network homogenization, increasing the risk of overfitting or underfitting. To address this, TLG effectively leverages heterogeneous information to capture richer semantic representations, thereby enhancing the model's robustness and generalization performance.

\begin{figure}[t]
    \centering
    \includegraphics[width=0.9\linewidth]{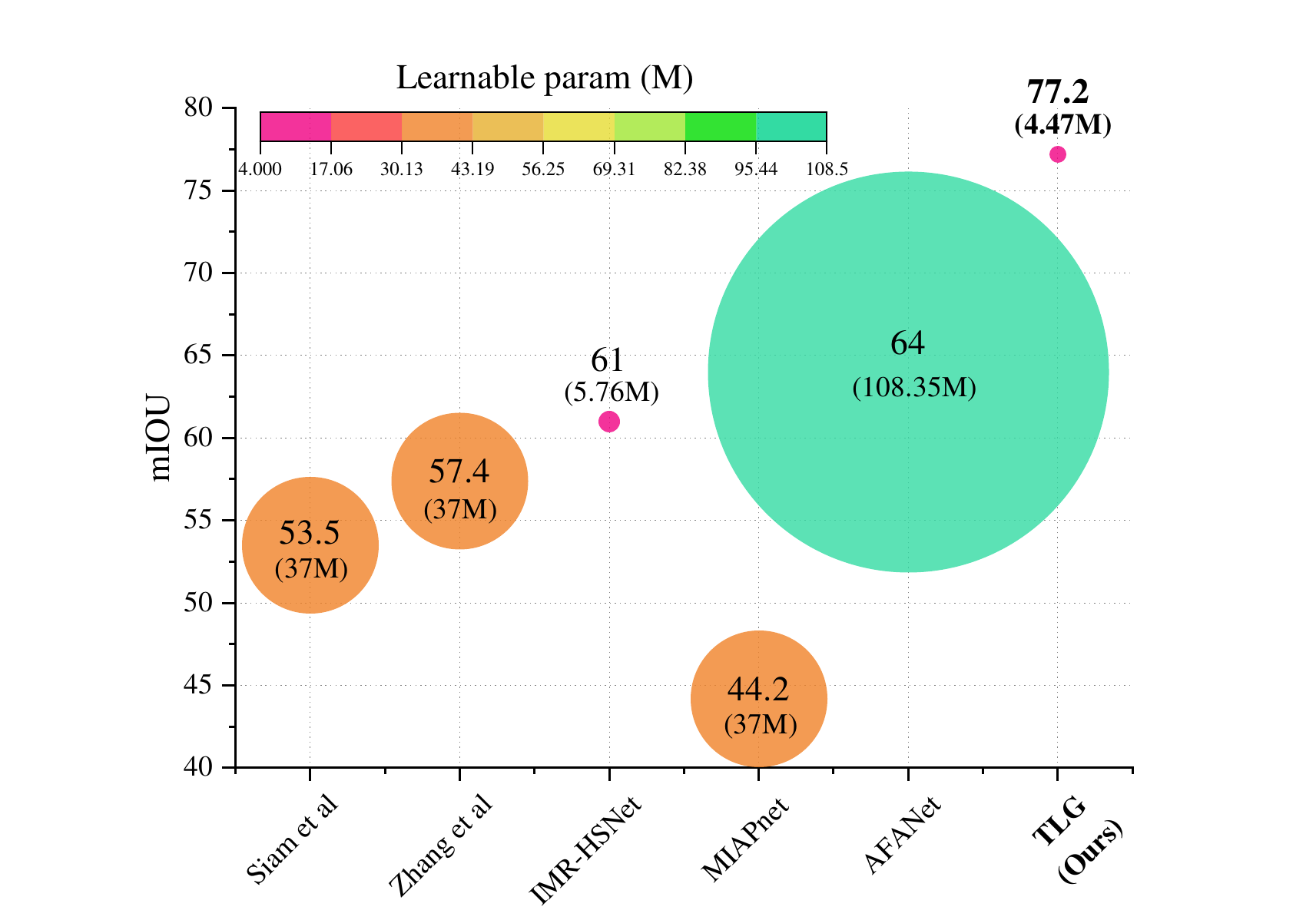}
    \caption{Analysis of the Lightweighting of Existing Models. The color and size represent different parameter scales.}
    \label{fig12}
    \vskip -0.15in
\end{figure}

\subsection{Weakly vs. Fully-supervised}
In this study, TLG utilizes the most challenging image-level labels. Compared to pixel-level labels, image-level labels only provide the overall category of each image, lacking precise pixel-level location information. Therefore, this imposes higher demands on the model's learning capability and robustness. As shown in Tables \ref{table1} and \ref{table2}, TLG outperforms the current state-of-the-art fully-supervised FSS models (with the same backbone). For instance, on the Pascal-5\textsuperscript{i} dataset, TLG achieves a performance of 77.2, compared to VRP-SAM's 71.9 and PGMA-Net's 74.1. To the best of our knowledge, this is the first time a weakly-supervised model has surpassed a fully-supervised one. This demonstrates TLG's ability to extract latent information from image-level labels, emphasizing its potential in few-shot semantic segmentation and the future of weakly-supervised learning.

\section{Lightweighting Analysis}
In this subsection, we conduct a comprehensive analysis of the parameter counts and performance of existing models. The learnable parameters for IMR-HSNet and AFANet are derived from experimental replication, while for Siam et al., Zhang et al., and MIAPNet, whose models are not open-source, we use the average parameter count from our replication experiments and TLG. The results demonstrate that TLG (4.47M) with only 1/24 of the parameters of existing state-of-the-art model AFANet (108.35M), TLG achieves a 13.2\% improvement.

Furthermore, considering the potential for lightweight edge deployment, we additionally evaluate inference-related metrics. As shown in Table \ref{table5}, we report the FLOPs and inference latency of TLG and the baseline model AFANet under different backbone and shot settings. For instance, with the ResNet50 backbone, TLG requires only 41.25G FLOPs for one-shot inference (70.4\% lower than AFANet’s 139.36G) and 206.24G for five-shot inference (70.4\% lower than AFANet’s 696.82G). Comparable reductions are observed on VGG16. In terms of latency, TLG achieves 36.72 ms for one-shot inference (50\% of AFANet’s 73.21 ms) and 174.38 ms for five-shot inference (49.7\% lower than AFANet’s 350.82 ms) on ResNet50, with consistently lower latency also observed on VGG16. This remarkable performance stems from the homology but heterogeneity network, which not only optimizes the architecture to substantially reduce parameters, but also enhances generalization through semantic complementarity.

\section{Limitations and Future Work}

Although the proposed heterogeneous layer selection strategy demonstrates strong performance, it is currently manually designed with fixed support-query layer assignments, which remain unchanged across inputs, limiting the model’s ability to dynamically exploit optimal hierarchical feature combinations and reducing its adaptability to different categories and instances. Potential avenues for future work include the following: 1) Learnable layer selection mechanisms, enabling data-dependent cross-layer interactions to replace fixed manual assignments. 2) Heterogeneous backbone architectures, introducing asymmetric feature specialization across branches. 3) Adaptive prompt generation, driven by foundation models, to replace fixed foreground–background prompts and further enhance cross-modal alignment and task adaptation capability.

\section{Conclusion}
In this study, we observed that the support-query pairs in meta-learning paradigms inherently exhibit homologous characteristics of the same categories and similar attributes. To address the issue of semantic over-homogenization caused by applying identical network architectures to such homologous data, we proposed a simple yet effective heterogeneous meta-learning framework, TLG. Extensive experiments and rigorous ablation studies across visual and linguistic multimodal paradigms demonstrate TLG's superior performance across multiple datasets. To the best of our knowledge, TLG is the first model to surpass fully supervised counterparts under the same backbone using only weak supervision. Beyond being a technical solution, TLG embodies a novel network design philosophy. We sincerely anticipate that this concept of heterogeneous architecture will inspire researchers and practitioners in addressing challenges within complex, diverse data ecosystems.

\bibliographystyle{IEEEtran}
\bibliography{reference} %

\vspace{11pt}
\begin{IEEEbiography}[{\includegraphics[width=1in,height=1.25in,clip,keepaspectratio]{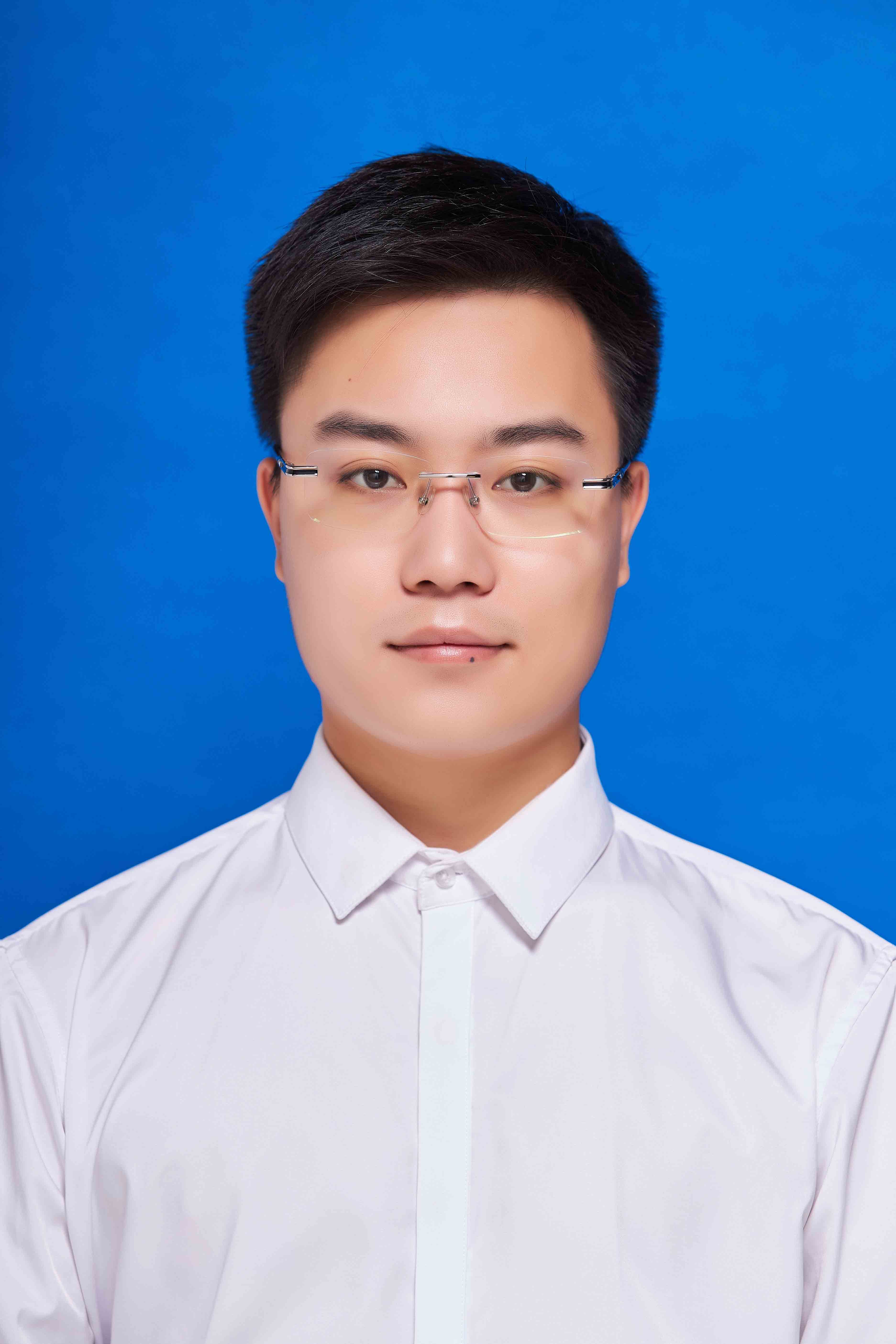}}] {Jiaqi Ma} received the B.S. degree in Harbin University of Commerce, Harbin, China, in 2022. He is currently pursuing the Ph.D. degree with the School of Computer Science and Engineering, Nanjing University of Science and Technology, China. His research interests include bioinformatics, computer vision, semantic segmentation, and multimodal model post-training.
\end{IEEEbiography}

\begin{IEEEbiography}[{\includegraphics[width=1in,height=1.25in,clip,keepaspectratio]{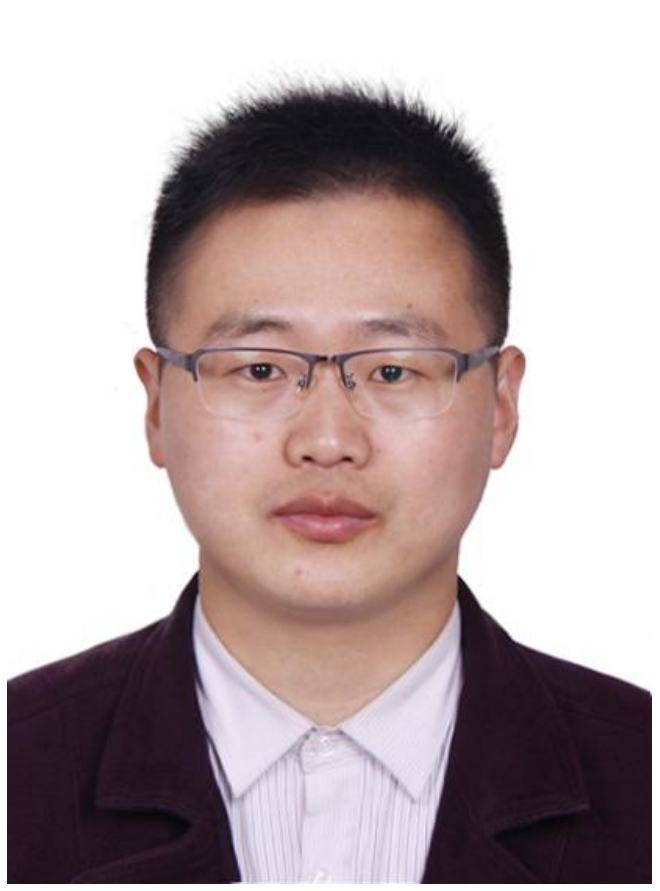}}] {Guo-Sen Xie} received the Ph.D. degree from the National Laboratory of Pattern Recognition, Institute of Automation, Chinese Academy of Sciences, Beijing, China, in 2016. He is a Professor at the School of Computer Science and Engineering, Nanjing University of Science and Technology, China. He received the Best Student Paper Award of MM 2016. He is an Associate Editor of IEEE T-IP and Pattern Recognition Journals, and Area Chairs of several international conferences, such as ICLR. His research interests include computer vision and machine learning.

\end{IEEEbiography}

\begin{IEEEbiography}[{\includegraphics[width=1in,height=1.25in,clip,keepaspectratio]{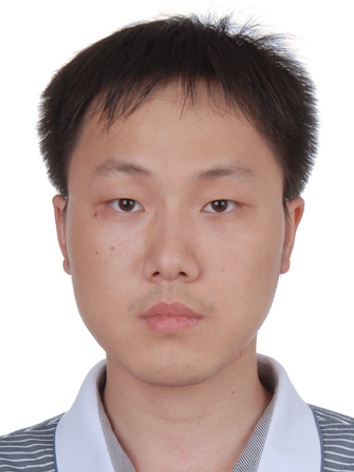}}] {Fang Zhao} received the Ph.D. degree from the National Laboratory of Pattern Recognition, Institute of Automation, Chinese Academy of Sciences, Beijing, China, in 2015. From October 2021 to June 2023, he worked as a Senior Researcher with the Tencent AI Lab, Shenzhen, China. He is currently an Associate Professor with the School of Intelligence Science and Technology, Nanjing University, Suzhou, China. His research interests include computer vision and machine learning.
\end{IEEEbiography}

\begin{IEEEbiography}[{\includegraphics[width=1in,height=1.25in,clip,keepaspectratio]{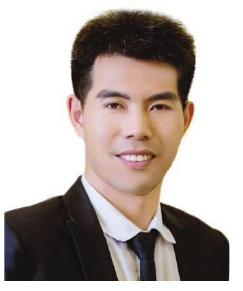}}] {Zechao Li} (Senior Member, IEEE) received the B.E. degree from the University of Science and Technology of China in 2008 and the Ph.D. degree from the National Laboratory of Pattern Recognition, Institute of Automation, Chinese Academy of Sciences, in 2013. He is currently a Professor at the Nanjing University of Science and Technology. His research interests include big media analysis and computer vision. He received the Best Paper Award at ACM Multimedia Asia 2020 and 2024. He is currently serving as an associate editor for IEEE TPAMI, IEEE TNNLS, IEEE TMM and IEEE TCSVT.
\end{IEEEbiography}

\vfill
\end{document}